# Using Learning Progressions to Guide AI Feedback for Science Learning

Xin Xia[1[0009-0009-1717-8511]], Nejla Yuruk [2[0000-0001-9240-750X]], Yun Wang[1[0009-0004-6611-0752]] and Xiaoming Zhai[1[0000-0003-4519-1931]]

[1] University of Georgia, Athens GA 30605, USA
[2] Gazi University, Ankara, 06560, Turkiye
xx86245@uga.com

**Abstract.** Generative artificial intelligence (AI) offers scalable support for formative feedback, yet most AI-generated feedback relies on task-specific rubrics authored by domain experts. While effective, rubric authoring is time-consuming and limits scalability across instructional contexts. Learning progressions (LP) provide a theoretically grounded representation of students' developing understanding and may offer an alternative solution. This study examines whether an LP–driven rubric generation pipeline can produce AI-generated feedback comparable in quality to feedback guided by expert-authored task rubrics. We analyzed AI-generated feedback for written scientific explanations produced by 207 middle school students in a chemistry task. Two pipelines were compared: (a) feedback guided by a human expert–designed, task-specific rubric, and (b) feedback guided by a task-specific rubric automatically derived from a learning progression prior to grading and feedback generation. Two human coders evaluated feedback quality using a multi-dimensional rubric assessing *Clarity*, *Accuracy*, *Relevance*, *Engagement and Motivation*, and *Reflectiveness* (10 sub-dimensions). Inter-rater reliability was high, with percent agreement ranging from 89% to 100% and Cohen's κ values for estimable dimensions (κ = .66–.88). Paired t-tests revealed no statistically significant differences between the two pipelines for *Clarity* ($t_1 = 0.00$, $p_1 = 1.000$; $t_2 = 0.84$, $p_2 = .399$), *Relevance* ($t_1 = 0.28$, $p_1 = .782$; $t_2 = -0.58$, $p_2 = .565$), *Engagement and Motivation* ($t_1 = 0.50$, $p_1 = .618$; $t_2 = -0.58$, $p_2 = .565$), or *Reflectiveness* ($t = -0.45$, $p = .656$). These findings suggest that the LP–driven rubric pipeline can serve as an alternative solution.

## 1 Introduction

Providing timely, high-quality formative feedback is essential in science education, particularly for open-ended explanation tasks that require students to articulate and justify their reasoning [1, 2]. For instance, in middle school chemistry, tasks such as ex-plaining gas properties and behaviors demand not only factual recall but also conceptual understanding of characteristic properties, evidence-based reasoning, and appropriate scientific language. Although these tasks are instructionally valuable, the effort required to generate individualized, pedagogically meaningful feedback places substantial demands on teachers' time and expertise, posing ongoing challenges for educators [3].

Recent advances in Artificial Intelligence, specifically large language models (LLMs), have evoked the potential in AI-generated feedback as a scalable approach to formative assessment. Prior work suggests that AI systems can generate feedback that is coherent, responsive, and comparable in overall quality to human feedback [4, 5]. However, research has consistently shown that the effectiveness of AI feedback depends not only on the underlying model, but also on how pedagogical structure is represented and communicated to the system [6, 7]. In practice, AI feedback systems rely on explicit human-authored artifacts, most notably rubrics, to guide what the system attends to and how it responds to student work [8, 9].

Rubrics play a critical role in both human and AI-generated feedback. By making evaluation criteria explicit, rubrics help align feedback with learning goals, support consistency, and foreground key conceptual elements and common misconceptions [10, 11]. In AI-generated feedback systems, rubrics serve as pedagogical components that constrain and scaffold the LLM model's reasoning, shaping both grading decisions and feedback content. Consequently, existing systems depend on task-specific rubrics authored by domain experts to ensure feedback quality and alignment. While effective, this reliance introduces a significant scalability challenge: developing high-quality rubrics for each new task, topic, or curriculum unit is time-intensive and requires specialized expertise, limiting the feasibility of deploying AI feedback broadly across classrooms and domains.

This challenge raises a fundamental design question for formative feedback-generating systems: Are human-authored, task-specific rubrics a necessary condition for LLMs generating high-quality feedback? If so, the scalability of AI feedback systems may remain constrained by human labor. If not, alternative forms of pedagogical structure may enable more sustainable and generalizable feedback pipelines.

Learning progressions (LP) offer a theoretically grounded pedagogical foundation for addressing this challenge through developing scalable alternatives to task-specific rubric design. Learning progressions describe empirically informed pathways through which students' understanding of core ideas typically develops over time, including intermediate conceptions and common misconceptions [12, 13]. Building on this perspective, the present study examines whether rubrics derived from learning progressions with task content can support AI-generated feedback of comparable quality to feedback guided by expert-authored, task-specific rubrics. Specifically, the study addresses the following research questions:

RQ1: What is the quality of AI-generated formative feedback guided by a learning progression-driven rubric?

RQ2: How does the quality of learning-progression-driven feedback compare to feedback guided by expert-authored rubric-driven feedback across specific dimensions, including *Clarity*, *Accuracy*, *Relevance*, *Engagement and Motivation*, and *Reflectiveness* guidance?

## 2 Literature Review

### 2.1 Feedback in Science Education

Providing timely, high-quality feedback is widely recognized as a critical factor in supporting student learning across disciplines. In science education, formative feedback plays a particularly important role because students are frequently asked to construct explanations, interpret data, and justify claims using evidence and scientific principles rather than selecting predefined answers [14, 15]. Effective feedback in these contexts helps students understand what counts as a high-quality scientific explanation, recognize gaps or weaknesses in their reasoning, and identify concrete next steps for improvement. However, providing timely, individualized feedback for such tasks remains a persistent challenge in classroom settings due to constraints on teachers' time and resources.

To address these scalability challenges, researchers have increasingly explored automated and AI-supported approaches to feedback and guidance. Early work in intelligent tutoring systems and adaptive learning environments demonstrated that automated guidance can support learning when it is carefully aligned with instructional goals and learner needs [7, 16]. More recently, advances in generative AI and large language models (LLMs) have expanded the scope of AI-generated feedback to include open-ended student work, such as written scientific explanations.

Emerging studies suggest that LLMs can generate feedback that is coherent, context-aware, and responsive to student input when provided with appropriate instructional structure [6, 17]. In science education specifically, recent research has examined the use of LLMs for automated scoring and feedback on constructed-response tasks, highlighting both the potential for rapid formative support and the importance of constraining AI outputs using pedagogically meaningful criteria [18, 19]. However, prior work cautions that when generative-AI feedback is not anchored in explicit evaluative criteria, such as detailed rubrics, exemplar responses, or other task-specific constraints, model outputs may drift toward overly generic or abstract comments and may not reliably align with the actual learning goals or evidence in students' work [20, 21]

Current AI feedback systems heavily rely on human-authored, task-specific rubrics to structure feedback generation and ensure alignment with disciplinary expectations. While effective, this approach raises concerns about scalability, particularly in middle school science contexts where explanation tasks vary widely across topics and phenomena. This tension motivates the exploration of alternative pedagogical foundations, such as learning progressions, that may support high-quality, developmentally aligned AI-generated feedback while reducing reliance on task-by-task human rubric authoring.

## 2.2 The Role of Rubrics in AI Feedback Systems

Rubrics are widely used in education to clarify expectations, guide assessment, and support consistent and meaningful feedback [10]. When AI systems are used to evaluate constructed responses and provide formative feedback, rubrics become especially important because they translate disciplinary standards (e.g., scientific explanation, evidence use, terminology precision) into concrete criteria that can be applied systematically at scale. In this sense, rubrics serve as a bridge between disciplinary expertise and automated decision-making, enabling feedback that is interpretable and aligned with instructional intent.

A large body of automated scoring research has shown that rubric design directly shapes the validity and reliability of automated scoring, and the quality of feedback derived from those scores. For example, research on automated scoring of constructed-response science items emphasizes that complex rubrics can be scored automatically with promising reliability, but that performance depends heavily on assessment and rubric design choices [22]. More recently, with the rise of large language models (LLMs), building on auto-scoring research, rubrics have been used to constrain and shape feedback generation. Several emerging systems guide LLM feedback using expert-defined criteria or rubric specifications to improve alignment and reduce generic output. For instance, [23] describes an LLM-assisted feedback tool in which instructor-defined criteria are used to guide automated responses to open-ended questions, emphasizing that criteria-based scaffolding supports more targeted and useful feedback. At the same time, empirical comparisons caution that generative AI feedback may fall short of human feedback on criteria-based qualities, such as prioritizing key issues and providing clear revision directions, when prompts and evaluative criteria are insufficiently specified [21].

## 2.3 Learning Progressions as Pedagogical Foundation

Learning progressions (LP) offer a theoretically grounded alternative for structuring assessment and feedback [24, 25]. LP describes empirically informed pathways through which students' understanding of core ideas typically develops, including intermediate conceptions and common misconceptions [12, 13]. In science education, LP has been used to guide curriculum design, assessment development, and instructional decision-making by articulating what it means to make progress toward disciplinary understanding.

Prior research has demonstrated the value of learning progressions as a foundation for assessment and automated analysis. LP-aligned assessments have been shown to improve the interpretability of student responses by linking observable performance features to underlying levels of understanding [19]. LP-aligned rubrics have been shown to support machine scoring that aligns with human judgments while providing interpretable indicators of student understanding. Prior research has leveraged learning progressions to support automated scoring and diagnostic classification [19, 26, 27]. For example, LP-aligned rubrics have been used to map student responses to developmental levels and to guide machine scoring of scientific explanations. However, most of this work has focused on classification accuracy or score validity, rather than on the quality of feedback generated for learners. As a result, it remains unclear whether LPs can effectively support the generation of high-quality, pedagogically meaningful feedback when used as the basis for automated rubric construction.

The present study addresses this gap by comparing two AI feedback generation pipelines in a middle school chemistry context involving a gas properties explanation task. By comparing these two pipelines using a within-subjects design, this work provides empirical evidence to address whether learning progressions can serve as a general pedagogical foundation that supports high-quality AI-generated feedback without relying on task-specific expert rubric authoring.

## 3 Methodology

### 3.1 Context and Datasets

The assessment task examined in this study was an open-ended, NGSS-aligned middle school chemistry task focused on students' understanding of gas properties. Students were presented with data from an investigation involving four gas samples, including information about flammability, volume, and density, and were asked to construct a written explanation identifying which gases could be the same based on characteristic properties and to justify their reasoning using evidence from the data table. The task was designed to elicit students' scientific explanations based on data interpretation, focusing on students' ability to analyze and interpret experimental data to determine whether substances are the same based on characteristic properties (Council et al., 2013).

This task was selected from the Next Generation Science Assessment task set [28] and falls within the physical sciences domain of Matter and Its Characteristics. The task was intentionally open-ended to encourage students to articulate reasoning, connect evidence to claims, and use appropriate scientific terminology. As such, it provides a rich context for examining formative feedback quality, particularly feedback that supports explanation construction rather than simply evaluating correctness, as Figure 1 shows below.

A total of 207 student responses used in this study were randomly drawn from a larger dataset of 1,200 middle school student responses collected through an NGSS-aligned online assessment system [29]. Student data were anonymized to protect privacy; therefore, no demographic information was available. Based on the geographic distribution of participating teachers, the dataset represents a broad cross-section of U.S. middle school students. The selected responses were used exclusively for generating and evaluating AI-produced formative feedback. The focus is on how different AI feedback generation pipelines respond to students' explanations and on the quality of that feedback across *Clarity*, *Accuracy*, *Relevance*, *Engagement and Motivation*, and *Reflectiveness* dimensions.

Gas filled balloons (ID#: 034.02-c01)

Tap text to listen

Alice did an experiment that caused four balloons to fill with gas, as shown in the figure to the right. Alice tested the flammability of each gas. She also measured the volume and mass of each gas to calculate the density. The tests and measures all occurred under the same conditions. The data is in Table 1.

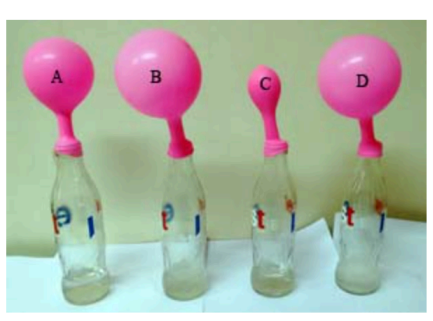

Table 1. Data of four gases in the balloons.

| Sample | Flammability | Density | Volume |
| --- | --- | --- | --- |
| Gas A | Yes | 0.089 g/L | 180 $cm^3$ |
| Gas B | No | 1.422 g/L | 270 $cm^3$ |
| Gas C | No | 1.981 g/L | 35 $cm^3$ |
| Gas D | Yes | 0.089 g/L | 269 $cm^3$ |

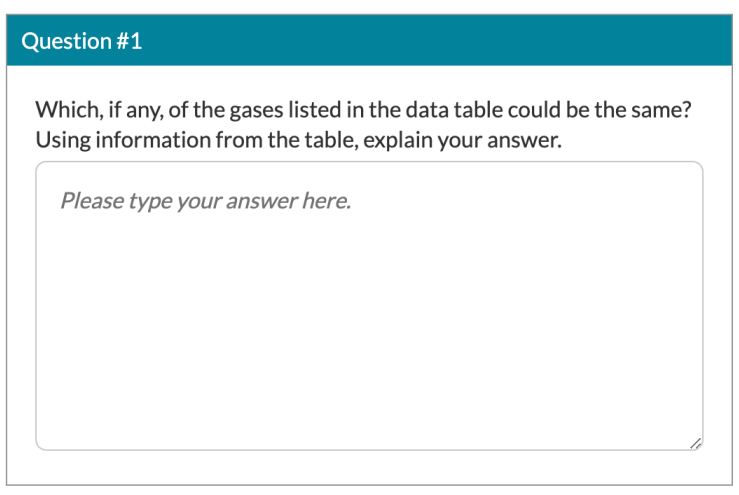

**Fig. 1.** The Gas-Filled Balloons Task

### 3.2 Study Design

We employed a within-subjects design to compare two AI-generated feedback pipelines. Each student response received feedback generated by both pipelines, allowing for direct comparison of feedback quality while both pipelines used the same student responses, task context, feedback structure, and large language model (GPT-5.1; temperature = 0). The learning progression is drawn from [30], including four levels (as Figure 2 shows below). The human-expert Rubric[27, 31] is listed in Figure 3.

The two feedback pipelines differed in how the pedagogical structure was provided to the AI system:

1. Expert-Rubric Pipeline: Feedback was generated using a human-authored, task-specific analytic rubric designed by science education experts. The rubric specified key conceptual criteria, common misconceptions, and performance expectations for the task.
2. Learning-Progression Pipeline: Feedback was generated using a task-specific rubric automatically derived from a five-level learning progression describing students' developing understanding of gas properties. In this pipeline, the AI first instantiated rubric criteria based on the learning progression and then used the generated rubric to evaluate student responses and produce feedback.

All other components, including task prompts, feedback templates, and the underlying language model, were held constant across conditions.

Feedback structure includes five steps: a) briefly acknowledge the student's answer, b) If students answer is not correct or partially correct, make the limitation visible or clarify and strengthen the missing key scientific idea, c) provide task-relevant next steps to improve the answer, d) prompts students to reflect on or justify their thinking, e) keep the feedback brief and clear.

**Level 4** Student constructs a complete evidence-based explanation

**Level 3** Student makes a claim and backs it up with sufficient and appropriate evidence but does not use reasoning to tie the two together

**Level 2** Student makes a claim and backs it up with appropriate but insufficient (partial) evidence

**Level 1** Student makes a claim with either no evidence or with inappropriate evidence

**Fig. 2.** Learning Progression for Evidence-based Explanations

| ID | Perspective | Description |
|---|---|---|
| E1 | SEP+DCI | Student states that Gas A and D could be the same substance. |
| E2 | SEP+CCC | Student describes the pattern (comparing data in different columns) in the table flammability data of Gas A and Gas D as the same. |
| E3 | SEP+CCC | Student describes the pattern (comparing data in different columns) in density data of Gas A and Gas D, which is the same in the table. |
| E4 | DCI | Student indicate flammability is one characteristic of identifying substances. |
| E5 | DCI | Student indicate density is one characteristic of identifying substances. |

**Fig. 3.** Learning Progression for Evidence-based Explanations

## 3.3 AI Feedback Generation

Feedback in both pipelines was generated using the same large language model (GPT-5.1). For each student response, the model was prompted to (a) evaluate the response according to the provided rubric and (b) generate formative feedback aligned with the evaluation. Feedback was designed to be developmentally appropriate, supportive in tone, and focused on helping students improve their scientific explanations.

To ensure comparability, both pipelines used identical feedback prompts and output constraints, differing only in the source of the rubric (human-authored versus learning-progression). This design isolates the effect of rubric origin on feedback quality.

## 3.4 Instrument

Feedback quality was evaluated using a multi-dimensional analytic rubric adapted and modified from the instrument developed by Field et al. (2025) [32]. The rubric assessed five dimensions of feedback quality, including *Clarity*, *Accuracy*, *Relevance*, *Engagement and Motivation*, and *Reflectiveness*. Each dimension was scored on a 3-point scale (0–2), yielding a maximum total feedback quality score of 20 (as Table 1 shows below).

**Table 1.** Feedback Quality Rubric.

| Dimension | Sub-dimension | Description |
|---|---|---|
| Clarity | Language | Language is accessible for the target age group. |
| | Structure | Feedback is brief, focused, and well-structured. |
| Accuracy | Correctness | Feedback contains no scientific misconceptions or factual errors. |
| | Terminology | Scientific terminology is used correctly and precisely. |
| Relevance | Responsiveness | Feedback is grounded in what the student actually said. |
| | Alignment | Feedback aligns with the goals and requirements of the task. |
| Engagement & Motivation | Tone | Feedback uses supportive and motivating language. |
| | Task | Feedback encourages interaction with the task. |
| Reflectiveness | Actionable | Feedback provides clear guidance for improvement when appropriate. |
| | Prompting | Feedback prompts students to reflect on or justify their thinking. |

### 3.5 Coding Procedure and Inter-Rater Reliability

Two human coders were trained using a shared codebook. Firstly, they independently coded an initial set of 20 AI-generated feedback, including rubric-driven and LP-driven feedback. The coders then met to discuss all discrepancies and refine their shared understanding of the rubric, revising interpretations as needed until full consensus (100% agreement) was reached on this calibration set.

Following the calibration process, the coders independently double-coded a randomly selected 20% of the full dataset to assess inter-rater reliability. Inter-rater reliability was calculated using both percent agreement and Cohen's Kappa(κ) for rubric dimensions with sufficient variability [33, 34]. Percent agreement across dimensions ranged from 89% to 100% (as Table 2 shows below). Cohen's κ values indicated moderate to strong agreement (κ = .66–.88). For several dimensions, κ could not be computed due to its sensitivity to the prevalence of agreement and limited score variance. After establishing acceptable reliability, the remaining feedback responses were randomly and equally divided between the two coders for independent coding [35].

**Table 2.** Inter-rater Reliability.

| Rubric Dimension | Percent Agreement (%) | Cohen's κ | Interpretation |
|---|---|---|---|
| Clarity – Language | 98.8 | 0.66 | Moderate |
| Clarity – Structure | 97.6 | 0.84 | Strong |
| Accuracy – Correctness | 98.8 | — | Not estimable |
| Accuracy – Terminology | 100.0 | — | Not estimable |
| Relevance – Responsiveness | 98.8 | — | Not estimable |
| Relevance – Alignment | 98.8 | — | Not estimable |
| Engagement – Tone | 89.0 | 0.82 | Strong |
| Engagement – Task | 100.0 | — | Not estimable |
| Reflectiveness – Actionable | 97.6 | −.01 | None |
| Reflectiveness – Prompting | 91.5 | 0.88 | Strong |

### 3.6 Data Analysis

Descriptive statistics were calculated for each feedback sub-dimension score under both rubric-driven and LP-driven approaches. To examine whether the two feedback-generating methods differed systematically, paired-samples t-tests were conducted for each sub-dimension. Statistical analyses were conducted using *R*, with an alpha level of .05 applied for all tests.

## 4 Results

### 4.1 Quality of the AI-generated Feedback

Across all sub-dimensions, mean scores were high for both pipelines ($M \geq 1.61$ on a 0–2 scale), indicating consistently strong feedback quality regardless of the rubric. For *Clarity*, mean scores for the *Language* sub-dimension reached the maximum value ($M = 2.00$) in both pipelines, indicating consistently age-appropriate and comprehensible feedback. Structure scores were similarly high (Rubric-driven: $M = 1.87$, $SD = 0.33$; LP-driven: $M = 1.85$, SD = 0.36).

For *Accuracy*, both scientific *Correctness* and *Terminology* precision achieved full mean scores ($M = 2.00$) with no observed variance across conditions, indicating uniformly accurate use of scientific content and terminology in both pipelines. Both Rubric-driven and LP-driven feedback contained no factual or conceptual errors and used scientific terminology correctly and precisely.

For *Relevance*, both *Responsiveness* to the student's answer (Rubric-driven: $M = 1.97$, $SD = 0.18$; LP-driven: $M = 1.96$, $SD = 0.19$) and *Alignment* with the task (Rubric-driven: $M = 1.99$, $SD = 0.10$; LP-driven: $M = 2.00$, $SD = 0.07$) were near full scores in both conditions. The result indicated that feedback was consistently grounded in students' written responses and aligned with the goals and requirements of the task, with only rare instances of loosely aligned feedback.

For *Engagement and Motivation*, mean scores were high for both encouraging *Tone* (Rubric-driven: $M = 1.91$, $SD = 0.35$; LP-driven: $M = 1.90$, $SD = 0.37$), indicating that feedback frequently used supportive and motivating language, and Task engagement (Rubric-driven: $M = 1.99$, $SD = 0.10$; LP-driven: $M = 2.00$, $SD = 0.07$), meaning feedback almost always encouraged students to engage further with the task or their reasoning.

Finally, for *Reflectiveness*, *Actionable* sub-dimension showed near-ceiling means (Rubric-driven: $M = 1.99$, $SD = 0.14$; *LP*: $M = 2.00$, $SD = 0.07$), whereas *Reflectiveness Prompting* exhibited comparatively lower mean scores (Rubric-driven: $M = 1.61$, $SD = 0.56$; LP: $M = 1.53$, $SD = 0.59$), indicating greater variability in this aspect of feedback quality. The results indicated that the two sets of feedback provided clear and appropriate guidance for improvement, and most feedback instances included reflective questions or prompts; some cases provided limited or no explicit opportunities for reflection, resulting in greater variability on this sub-dimension.

This overall pattern supports the conclusion that both Rubric-driven and LP-driven pipelines are able to generate high-quality feedback. Two examples of the full-score feedback of Rubric-driven and LP-driven are shown below (Figure 4):

Students answer

"*Gas A and B could be the same because they both share the same properties.*"

Rubric-driven Feedback

"*You noticed that some gases might share properties, which is a good start, but Gas A and Gas B actually do not have the same flammability or density in the table.*

*Look back at the table and ask yourself: which gases have both the same flammability result and the exact same density number, and why are those two properties more useful for identifying a substance than volume?*"

LP- Feedback:

"*You're on the right track thinking about 'properties,' but A and B actually don't share the same test results in the table.*

*Look carefully at the density numbers and the flammability "yes/no" for A and B: which of those are the same and which are different, and what does that tell you about whether they could be the same gas?*"

**Fig. 4.** Examples of Feedback

### 4.2 Comparison of Feedback Quality Between the Two Pipelines

To test whether the feedback quality difference between the Rubric–driven and LP-driven pipelines, we conducted paired-samples t-tests for each feedback sub-dimension (Table 3). Because each of the 207 student responses received feedback from both pipelines, the paired design compared within-response differences while controlling student response content. Degrees of freedom were 206 for all estimable tests.

Overall, the paired t-tests indicated no statistically significant differences between the two pipelines across any feedback sub-dimension (all $p > .05$). For *Clarity*, no differences were detected for the *Language* accessibility ($t = 0.00$, $df = 206$, $p = 1.000$) or *Structure* ($t = 0.84$, $df = 206$, $p = .399$). For *Relevance*, the sub-dimension of *Responsiveness* to the student answer did not differ across pipelines ($t = 0.28$, $df = 206$, $p = .782$), nor did the *Alignment* with task requirements ($t = -0.58$, $df = 206$, $p = .565$). Similarly, for *Engagement and Motivation*, no significant differences were observed for encouraging *Tone* ($t = 0.50$, $df = 206$, $p = .618$) or *Task* engagement ($t = -0.58$, $df = 206$, $p = .565$). For *Reflectiveness*, *Actionable* also showed no difference ($t = -0.45$, $df = 206$, $p = .656$). Reflective *Prompting* yielded the largest observed test statistic but remained non-significant ($t = 1.72$, $df = 206$, $p = .088$).

For the two *Accuracy* sub-dimensions, scientific *Correctness* and *Terminology* precision, paired t-tests were not estimable because all feedback instances received the maximum score ($M = 2.00$) in both pipelines, resulting in zero variance and no within-pair differences.

Taken together, these inferential results indicate that, across all evaluable sub-dimensions, feedback quality did not differ statistically between the Rubric–driven and LP-driven pipelines under the conditions of this study.

**Table 3.** Feedback Evaluation T-test.

| Sub-Dimension | Rubric *M* (*SD*) | LP *M* (*SD*) | *t* | *df* | *p* |
|---|---|---|---|---|---|
| Clarity – Language | 2.00 (0.07) | 2.00 (0.07) | 0.00 | 206 | 1.000 |
| Clarity – Structure | 1.87 (0.33) | 1.85 (0.36) | 0.84 | 206 | 0.399 |
| Accuracy – Correctness | 2.00 (0.00) | 2.00 (0.00) | — | 206 | — |
| Accuracy – Terminology | 2.00 (0.00) | 2.00 (0.00) | — | 206 | — |
| Relevance – Responsiveness | 1.97 (0.18) | 1.96 (0.19) | 0.28 | 206 | 0.782 |
| Relevance – Alignment | 1.99 (0.10) | 2.00 (0.07) | −0.58 | 206 | 0.565 |
| Engagement – Tone | 1.91 (0.35) | 1.90 (0.37) | 0.50 | 206 | 0.618 |
| Engagement – Task | 1.99 (0.10) | 2.00 (0.07) | −0.58 | 206 | 0.565 |
| Reflectiveness – Actionable | 1.99 (0.14) | 2.00 (0.07) | −0.45 | 206 | 0.656 |
| Reflectiveness – Prompting | 1.61 (0.56) | 1.53 (0.59) | 1.72 | 206 | 0.088 |

# 5 Discussion

## 5.1 Feedback Quality

Across both pipelines, feedback quality was consistently high on the 0–2 rubric scale, with near-ceiling means for most sub-dimensions (e.g., *Clarity*, *Relevance*, *Engagement*) and perfect scores for scientific *Correctness* and *Terminology*. In other words, feedback consistently met the rubric's highest quality standards across evaluated dimensions, clear and age-appropriate phrasing, strong alignment to the student response and task, supportive tone, and actionable guidance. Importantly, *Accuracy* was consistently high in this dataset: coders did not identify scientific misconceptions or incorrect terminology in any of the 414 feedback instances, resulting in zero variance for the two accuracy sub-dimensions. This pattern suggests that, under a constrained prompt and a well-defined task context, the system can reliably produce scientifically sound feedback at scale.

This result is consistent with a broader line of work showing that automated feedback can be effective when it is tightly aligned to task goals and provides actionable guidance rather than generic comments. For example, a previous study conducted by Zhu et al. (2020)[36] has shown that automated feedback embedded in science argumentation tasks has been shown to shape students' revision behavior, especially when feedback is contextualized to the student's work rather than generic. Related work in science education conducted by Lee et al. (2019) [17] has also demonstrated the feasibility of automated text scoring with real-time, revision-oriented feedback for scientific arguments, emphasizing that carefully engineered feedback systems can support student improvement in authentic science practices. Although prior studies relied on human-expert–designed rubrics to guide automated evaluation, this study illustrates an alternative and generalizable approach to integrating LP-driven structures into automated feedback systems.

Recent studies that evaluate AI feedback using analytic rubrics increasingly highlight that the structure given to the model (criteria, constraints, context) strongly influences feedback quality. A large-scale study comparing AI-generated and human-authored formative feedback using an expert rubric found that AI feedback can reach strong quality on multiple pedagogical dimensions, though quality varies by dimension and context [37]. Similarly, work on criterion-based evaluation with detailed criteria shows that providing explicit criteria can improve the reliability and defensibility of automated judgments [38]. These patterns are consistent with our design: both pipelines supplied explicit evaluative, and both produced high-quality outputs.

### 5.2 Synthesis with rubric-based feedback research

A key contribution of this study is not just that the feedback was high quality, but that LP-derived rubric criteria performed comparably to expert-authored rubric criteria. This is important because the majority of rubric-based automated feedback work depends on carefully crafted, task-specific rubrics to ensure feedback is diagnostic, aligned, and actionable. In science education contexts, systems like automated text scoring with real-time feedback show that structured criteria can support learners' revision and argument quality[17]. At the same time, task-by-task rubric development remains a practical bottleneck for scaling AI feedback across diverse classroom tasks.

The findings suggest a synthesis: learning progressions can function as an alternative pedagogical foundation that enables systems to generate task-relevant criteria (rubric instantiation) without requiring experts to author a bespoke rubric for every new phenomenon. This aligns with recent LP with automated analysis work emphasizing that LP-aligned tools can support assessment and learning in STEM by anchoring automated interpretations in developmental descriptions of student thinking [27]. In design terms, these results support a "domain backbone" → "task instantiation" approach: experts invest in validating learning progressions, and AI systems instantiate task rubrics and feedback from that backbone.

## 6 Implementation and Limitation

The current study focused on a middle school chemistry explanation task involving characteristic properties of gases. Although this task represents a common type of NGSS-aligned explanation activity, the findings may not generalize to other disciplines (e.g., Mathematics, Language Arts). This approach also needs more empirical evidence on other science domains, grade levels, or task formats (e.g., modeling, data analysis across extended datasets, or multi-step investigations). Future work should examine whether learning-progression–driven feedback performs similarly across a broader range of scientific practices and content areas.

Secondly, this study evaluated feedback quality rather than student learning outcomes. While high-quality feedback is a necessary condition for supporting learning, the present analysis does not assess how students interpret, use, or act upon the feedback, nor whether feedback generated through different pipelines leads to differential learning gains. Prior research suggests that feedback uptake and revision behavior are influenced by classroom context and instructional integration; thus, future studies should investigate the impact of LP-driven feedback on student revision and learning over time.

**Acknowledgments.** This material is based upon work supported by the Institute of Education Sciences (No. R305C240010) and the National Science Foundation (No. 2101104). Any opinions, findings, and conclusions are those of the authors and do not reflect the views of the Institute of Education Sciences or the National Science Foundation.

## References

1. McNeill, K.L., Krajcik, J.S.: Supporting Grade 5-8 Students in Constructing Explanations in Science: The Claim, Evidence, and Reasoning Framework for Talk and Writing. Pearson (2011).
2. Shute, V.J.: Focus on Formative Feedback. Rev. Educ. Res. 78, 153–189 (2008). https://doi.org/10.3102/0034654307313795.
3. Hattie, J., Timperley, H.: The Power of Feedback. Rev. Educ. Res. 77, 81–112 (2007). https://doi.org/10.3102/003465430298487.
4. Abar, R.O., Pong, M., Som, R.: AI-Driven Feedback Systems for Formative Assessment: Toward Personalized and Real-Time Pedagogy. Al-Hijr J. Adulearn World. 4, 87–100 (2025). https://doi.org/10.55849/alhijr.v4i2.984.
5. Deepshikha, D.: A systematic review on the future of educational assessment: AI-driven grading and personalised feedback in higher education. Artif. Intell. Educ. 1–41 (2025). https://doi.org/10.1108/AIIE-03-2025-0036.
6. Bulut, O., Wongvorachan, T.: Feedback Generation through Artificial Intelligence. Open-Technology Educ. Soc. Scholarsh. Assoc. Conf. 2, 1–9 (2022). https://doi.org/10.18357/otessac.2022.2.1.125.
7. VanLehn, K.: The Relative Effectiveness of Human Tutoring, Intelligent Tutoring Systems, and Other Tutoring Systems. Educ. Psychol. 46, 197–221 (2011). https://doi.org/10.1080/00461520.2011.611369.
8. Lai, P., Lau, I., Pang, R.: Exploring the Efficacy of Rubric-Based AI Feedback in Enhancing Student Writing Outcomes. In: 2024 6th International Workshop on Artificial Intelligence and Education (WAIE). pp. 220–224 (2024). https://doi.org/10.1109/WAIE63876.2024.00047.
9. Pan, Y.: Leveraging generative AI powered rubric-indexed feedback as a formative assessment strategy for enhancing medical English education. Discov. Comput. 28, 284 (2025). https://doi.org/10.1007/s10791-025-09830-9.
10. Brookhart, S.M.: Appropriate Criteria: Key to Effective Rubrics. Front. Educ. 3, (2018). https://doi.org/10.3389/feduc.2018.00022.
11. Panadero, E., Andrade, H., Brookhart, S.: Fusing self-regulated learning and formative assessment: a roadmap of where we are, how we got here, and where we are going. Aust. Educ. Res. 45, 13–31 (2018). https://doi.org/10.1007/s13384-018-0258-y.
12. Alonzo, A.C., Steedle, J.T.: Developing and assessing a force and motion learning progression. Sci. Educ. 93, 389–421 (2009). https://doi.org/10.1002/sce.20303.
13. Duncan, R.G., Hmelo-Silver, C.E.: Learning progressions: Aligning curriculum, instruction, and assessment. J. Res. Sci. Teach. 46, 606–609 (2009). https://doi.org/10.1002/tea.20316.
14. Bell, B., Cowie, B.: The characteristics of formative assessment in science education. Sci. Educ. 85, 536–553 (2001). https://doi.org/10.1002/sce.1022.

15. Black, P.: Assessment and feedback in science education. Stud. Educ. Eval. 21, 257–279 (1995). https://doi.org/10.1016/0191-491X(95)00015-M.
16. Koedinger, K.R., D'Mello, S., McLaughlin, E.A., Pardos, Z.A., Rosé, C.P.: Data mining and education. WIREs Cogn. Sci. 6, 333–353 (2015). https://doi.org/10.1002/wcs.1350.
17. Lee, H.-S., Pallant, A., Pryputniewicz, S., Lord, T., Mulholland, M., Liu, O.L.: Automated text scoring and real-time adjustable feedback: Supporting revision of scientific arguments involving uncertainty. Sci. Educ. 103, 590–622 (2019). https://doi.org/10.1002/sce.21504.
18. Jescovitch, L.N., Scott, E.E., Cerchiara, J.A., Merrill, J., Urban-Lurain, M., Doherty, J.H., Haudek, K.C.: Comparison of Machine Learning Performance Using Analytic and Holistic Coding Approaches Across Constructed Response Assessments Aligned to a Science Learning Progression. J. Sci. Educ. Technol. 30, 150–167 (2021). https://doi.org/10.1007/s10956-020-09858-0.
19. Kaldaras, L., Haudek, K., Krajcik, J.: Employing automatic analysis tools aligned to learning progressions to assess knowledge application and support learning in STEM. Int. J. STEM Educ. 11, 57 (2024). https://doi.org/10.1186/s40594-024-00516-0.
20. Jukiewicz, M., Wyrwa, M.: Can ChatGPT Replace the Teacher in Assessment? A Review of Research on the Use of Large Language Models in Grading and Providing Feedback. Appl. Sci. 16, (2026). https://doi.org/10.3390/app16020680.
21. Steiss, J., Tate, T., Graham, S., Cruz, J., Hebert, M., Wang, J., Moon, Y., Tseng, W., Warschauer, M., Olson, C.B.: Comparing the quality of human and ChatGPT feedback of students' writing. Learn. Instr. 91, 101894 (2024). https://doi.org/10.1016/j.learninstruc.2024.101894.
22. Liu, O.L., Brew, C., Blackmore, J., Gerard, L., Madhok, J., Linn, M.C.: Automated Scoring of Constructed-Response Science Items: Prospects and Obstacles. Educ. Meas. Issues Pract. 33, 19–28 (2014). https://doi.org/10.1111/emip.12028.
23. Matelsky, J.K., Parodi, F., Liu, T., Lange, R.D., Kording, K.P.: A large language model-assisted education tool to provide feedback on open-ended responses, http://arxiv.org/abs/2308.02439, (2023). https://doi.org/10.48550/arXiv.2308.02439.
24. Council, N.R., Education, D. of B. and S.S. and, Education, B. on S., Standards, C. on a C.F. for N.K.-12 S.E.: A Framework for K-12 Science Education: Practices, Crosscutting Concepts, and Core Ideas. National Academies Press (2012).
25. Wilson, M.: Measuring progressions: Assessment structures underlying a learning progression. J. Res. Sci. Teach. 46, 716–730 (2009). https://doi.org/10.1002/tea.20318.
26. Jin, H., Lima, C., Wang, L.: Automated Scoring in Learning Progression-Based Assessment: A Comparison of Researcher and Machine Interpretations. Educ. Meas. Issues Pract. 44, 25–37 (2025). https://doi.org/10.1111/emip.70003.
27. Kaldaras, L., Haudek, K.C.: Validation of automated scoring for learning progression-aligned Next Generation Science Standards performance assessments. Front. Educ. 7, (2022). https://doi.org/10.3389/feduc.2022.968289.
28. Harris, C.J., Krajcik, J.S., Pellegrino, J.W.: Creating and using instructionally supportive assessments in NGSS classrooms. NSTA Press, National Science Teaching Association (2024).
29. Zhai, X., He, P., Krajcik, J.: Applying machine learning to automatically assess scientific models. J. Res. Sci. Teach. 59, 1765–1794 (2022). https://doi.org/10.1002/tea.21773.

30. Gotwals, A.W., Songer, N.B.: Validity Evidence for Learning Progression-Based Assessment Items That Fuse Core Disciplinary Ideas and Science Practices. J. Res. Sci. Teach. 50, 597–626 (2013). https://doi.org/10.1002/tea.21083.
31. He, P., Shin, N., Zhai, X., Krajcik, J.: A Design Framework for Integrating Artificial Intelligence to Support Teachers' Timely Use of Knowledge-in-Use Assessments. (2023). https://doi.org/10.13140/RG.2.2.19088.58881.
32. Watts, F.M., Liu, L., Ober, T.M., Song, Y., Valle, E.J.-D., Zhai, X., Wang, Y., Liu, N.: A Framework for Designing an AI Chatbot to Support Scientific Argumentation. Educ. Sci. 15, (2025). https://doi.org/10.3390/educsci15111507.
33. Cohen, J.: A Coefficient of Agreement for Nominal Scales. Educ. Psychol. Meas. 20, 37–46 (1960). https://doi.org/10.1177/001316446002000104.
34. McHugh, M.L.: Interrater reliability: the kappa statistic. Biochem. Medica. 22, 276–282 (2012).
35. Li, M., Gao, Q., Yu, T.: Kappa statistic considerations in evaluating inter-rater reliability between two raters: which, when and context matters. BMC Cancer. 23, 799 (2023). https://doi.org/10.1186/s12885-023-11325-z.
36. Zhu, M., Liu, O.L., Lee, H.-S.: The effect of automated feedback on revision behavior and learning gains in formative assessment of scientific argument writing. Comput. Educ. 143, 103668 (2020). https://doi.org/10.1016/j.compedu.2019.103668.
37. Nazaretsky, T., Gabbay, H., Käser, T.: Can students judge like experts? A large-scale study on the pedagogical quality of AI and human personalized formative feedback. Comput. Educ. Artif. Intell. 10, 100533 (2026). https://doi.org/10.1016/j.caeai.2025.100533.
38. Zhang, D.-W., Boey, M., Tan, Y.Y., Jia, A.H.S.: Evaluating large language models for criterion-based grading from agreement to consistency. NPJ Sci. Learn. 9, 79 (2024). https://doi.org/10.1038/s41539-024-00291-1.